 \documentclass[pmlr,twocolumn,10pt]{jmlr} 

\usepackage[utf8]{inputenc} 
\usepackage[T1]{fontenc}    
\usepackage{hyperref}       
\usepackage{url}            
\usepackage{booktabs}       
\usepackage{amsfonts}       
\usepackage{nicefrac}       
\usepackage{microtype}      
\usepackage{xcolor}         
\usepackage{graphicx}
\usepackage{enumitem}
\usepackage{amsmath}
\usepackage{multirow}
\usepackage{multicol}
\usepackage[normalem]{ulem}
\useunder{\uline}{\ul}{}
\usepackage{tablefootnote}
\usepackage[labelfont=bf]{caption}

\usepackage{verbatim}
\usepackage{etoolbox}
\usepackage{fancyhdr}
\usepackage{lipsum}

\usepackage{booktabs}
\usepackage[load-configurations=version-1]{siunitx} 

\newcommand{\equal}[1]{{\hypersetup{linkcolor=black}\thanks{#1}}}

\theorembodyfont{\upshape}
\theoremheaderfont{\scshape}
\theorempostheader{:}
\theoremsep{\newline}

\jmlrvolume{Index}
\jmlryear{2021}
\jmlrsubmitted{LEAVE UNSET}
\jmlrpublished{LEAVE UNSET}
\jmlrworkshop{Machine Learning for Health (ML4H) 2021} 

\title[Unifying Heterogeneous EHR Systems via Text-Based Code Embedding]{Unifying Heterogeneous Electronic Health Records Systems via Text-Based Code Embedding}

 \author{%
  \Name{Kyunghoon Hur}\equal{These authors contributed equally} \Email{pacesun@kaist.ac.kr}\\
  \Name{Jiyoung Lee}\footnotemark[1] \Email{jiyounglee0523@kaist.ac.kr}\\
  \Name{Jungwoo Oh} \Email{ojw0123@kaist.ac.kr}\\
  \addr Graduate School of AI, KAIST, South Korea
  \AND
  \Name{Wesley Price} \Email{wjprice@mit.edu}\\
  \addr MIT, USA
  \AND
  \Name{Younghak Kim} \Email{mdyhkim@amc.seoul.kr}\\
  \addr Asan Medical Center, University of Ulsan College of Medicine, South Korea
   \AND
   \Name{Edward Choi} \Email{edwardchoi@kaist.ac.kr}\\
   \addr Graduate School of AI, KAIST, South Korea
 }

\begin{document}

\setlength{\belowcaptionskip}{-12pt}

\maketitle

\begin{abstract}
    EHR systems lack a unified code system for representing medical concepts, which acts as a barrier for the deployment of deep learning models in large scale to multiple clinics and hospitals. To overcome this problem, we introduce Description-based Embedding, \textit{DescEmb}, a code-agnostic representation learning framework for EHR. DescEmb takes advantage of the flexibility of neural language understanding models to embed clinical events using their textual descriptions rather than directly mapping each event to a dedicated embedding. DescEmb outperformed traditional code-based embedding in extensive experiments, especially in a zero-shot transfer task (one hospital to another), and was able to train a single unified model for heterogeneous EHR datasets.
\end{abstract}


\vspace{-5mm}
\section{Introduction}
\label{sec:intro}
\vspace{-2mm}
Increased adoption of electronic health record (EHR) systems offers great potential for predictive models to improve healthcare quality. Some prior works use autoencoders \citep{miotto2016deep, che2015deep}, recurrent neural network \citep{choi2016doctor, choi2016retain, pham2016deepcare} and other deep learning architecture \citep{choi2016multi, choi2017gram} to perform prediction, interpretability or scalability. However, the heterogeneity of the code systems of EHR present barriers for deep learning model applications. Contemporary EHRs rely on data systems ranging from standardized codes (e.g. ICD9) to free-text entry. Previous approaches are based on learning the representations of these codes, an approach we refer to as "code-based embedding". However, this paradigm is not flexible to be transferred from one environment to another nor be trained on large EHR data where different EHR formats are used. Consequently, modern deep learning prediction models are missing out on the opportunity to be scaled up to multi-regional or even multi-national data. This challenge could be alleviated by mapping codes from one system to another, or by converting all EHR data to Common Data Model (e.g. OMOP, FHIR) \citep{rajkomar2018scalable}. However, mapping codes from one system to another requires significant human effort and domain knowledge and may not even be possible, depending on the code system at hand.

In this paper, we address this challenge by suggesting a code-agnostic  description-based Embedding, DescEmb. Based on the idea that each code has its description, DescEmb adopts aforementioned neural text encoder to convert medical code to contextualized embeddings, mapping medical codes of different formats to the same text embedding space. Figure \ref{Overview} in Appendix~\ref{supp:fig} visualizes our motivation. The best model of DescEmb demonstrates comparable performance to the best model of CodeEmb in the majority of cases, outperforming by an average of 2.6\%P AUPRC.
\vspace{-5mm}
\section{Methods}
\vspace{-3mm}
\subsection{Model Architecture}
\begin{figure}[!htbp]
  \includegraphics[width=\linewidth]{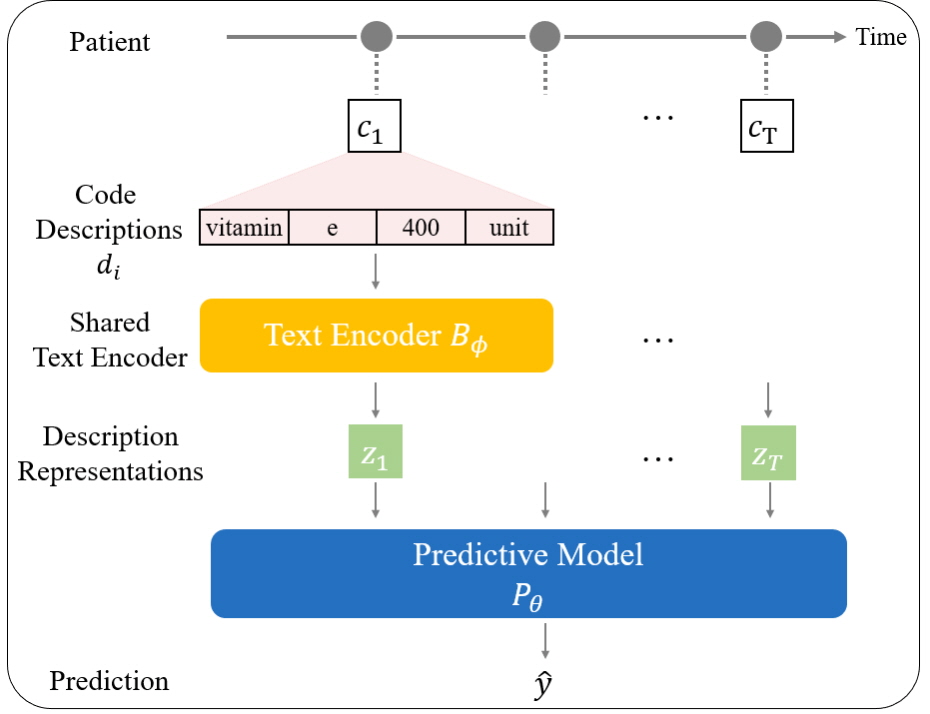}
  \caption{\label{Model} \textbf{DescEmb model framework.}}
\end{figure}

\vspace{-3mm}
A single patient can be seen as a series of medical events $(c_{1}, c_{2}, \ldots, c_{T})$ for $c_{i} \in \mathcal{C}$ where $\mathcal{C}$ denotes the set of all medical events such as diagnoses or prescriptions. Previous predictive models for EHR data use an embedding layer (or a lookup table) $\textit{E}_{\psi}$ which converts a single medical event $c_{i}$ to its vector representation $\textbf{\textit{c}}_{i} \in \mathbb{R}^a$ where $a$ is the dimension size. 

Instead of directly converting $c_1, \ldots, c_T$ to $\textbf{\textit{c}}_1, \ldots, \textbf{\textit{c}}_T$, DescEmb derives the latent representation of $c_{i}$, based on its text description $d_{i}$. We feed $d_{i}$ to the shared text encoder, $\textit{B}_{\phi}$, to obtain the description representations, $\textbf{\textit{z}}_{i} \in \mathbb{R}^{b}$ where $b$ is the output dimension. Repeating this for all events in the given patient, we can obtain a sequence of contextualized medical event representations $(\textbf{\textit{z}}_{1}, \textbf{\textit{z}}_{2}, \ldots, \textbf{\textit{z}}_{T})$, which is given to the prediction layer $\textit{P}_{\theta}$ (e.g. RNN) to make a prediction $\hat{y}$ (Fig \ref{Model}.) The entire process of DescEmb can be summarized as below, with comparison to CodeEmb.
\vspace{-7mm}

\begin{flalign}
& \mbox{Given a patient record $p = (c_1, c_2, \ldots, c_T$),} \nonumber && \\
& \quad \mbox{\textit{Code-based Embedding}:} \nonumber && \\
& \qquad \quad \textbf{\textit{c}}_i = \textit{E}_\psi(c_i) \nonumber && \\
& \qquad \quad \hat{y} = \textit{P}_{\theta}(\textbf{\textit{c}}_1, \textbf{\textit{c}}_2, \ldots, \textbf{\textit{c}}_T) && \\
& \quad \mbox{\textit{Description-based Embedding}:} \nonumber && \\
& \qquad \quad d_i = (w_{i,1}, w_{i,2}, \ldots, w_{i,n}) && \nonumber \\
& \qquad \quad \textbf{\textit{z}}_i = \textit{B}_{\phi}(d_i) && \\
& \qquad \quad \hat{y} = \textit{P}_{\theta}(\textbf{\textit{z}}_1, \textbf{\textit{z}}_2 \ldots, \textbf{\textit{z}}_T) &&  \nonumber
\end{flalign}

\vspace{-6mm}
\subsection{Value Embedding}
\vspace{-2mm}
Values incorporated in a code description provide informative features to the model. When using DescEmb, both the code description $d_{i}$ and the numeric values can be embedded with the text encoder $\textit{B}_{\phi}$. We split all numeric values into each digit, then aggregate with the code description as text named Digit-Split Value Aggregation(DSVA). Also, we add learnable Digit Place Embedding(DPE) to every digit token indicating its place value. Value Concatenated (VC) embeds code description through the text encoder, while values through additional Multi-Layer Perceptron (MLP). These two embeddings are finally concatenated and work as input of the predictive model.
\vspace{-5mm}
\begin{figure}[!htbp]
  \includegraphics[width=\linewidth]{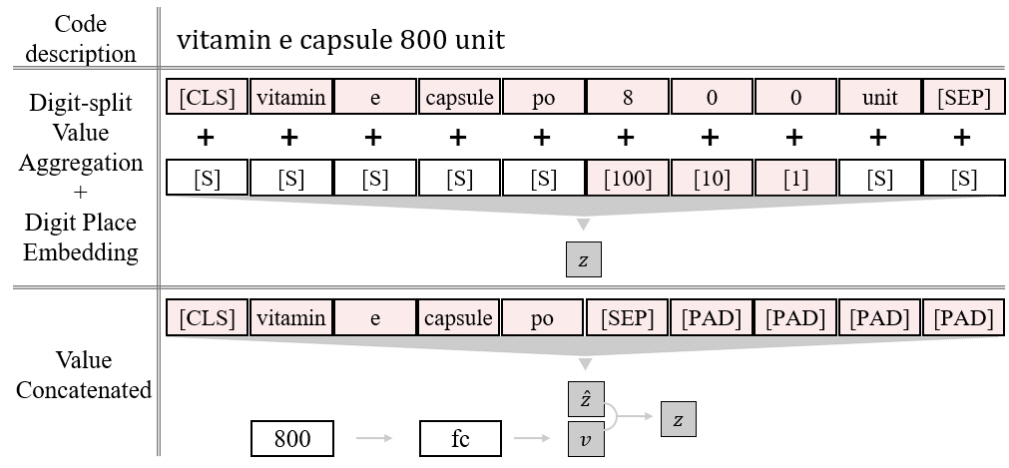}
  \caption{\label{value} \textbf{Methods of incorporating numeric values.}}
\end{figure}
\vspace{-6mm}
\section{Experiments and Results}


\begin{table*}[ht]
    \caption{\label{pooled} \textbf{AUPRC of the models on the five prediction tasks in the three scenarios: single domain learning, transfer learning, pooled learning.} We compared the AUPRC of code-based embedding model (CodeEmb), pre-trained BERT model (FT-BERT), RNN model (SC-RNN), and RNN model pre-trained on Masked Language Modeling (SC-RNN+MLM).
    Based on t-test, statistically meaningful increase and decrease against “Single” is marked with boldface and underline, respectively. }
    \centering
    \resizebox{\textwidth}{!}{\begin{tabular}{cc|ccc|ccc}
    \toprule
                                     &                        & \multirow{2}{*}{\begin{tabular}[c]{@{}c@{}}Single\\ MIMIC-III\end{tabular}} & \multirow{2}{*}{\begin{tabular}[c]{@{}c@{}}Transfer\\ eICU \\  → MIMIC-III\end{tabular}} & \multirow{2}{*}{\begin{tabular}[c]{@{}c@{}}Pooled \\  MIMIC-III\end{tabular}} & \multirow{2}{*}{\begin{tabular}[c]{@{}c@{}}Single\\ eICU\end{tabular}} & \multirow{2}{*}{\begin{tabular}[c]{@{}c@{}}Transfer\\ MIMIC-III \\ → eICU\end{tabular}} & \multirow{2}{*}{\begin{tabular}[c]{@{}c@{}}Pooled\\ eICU\end{tabular}} \\
                \multirow{2}{*}{Task}                                & \multirow{2}{*}{Model} &                                                                             &                                                                                          &                                                                               &                                                                        &                                                                                         &                                                                        \\
                                     &                        & \multicolumn{1}{l}{}                                                        & \multicolumn{1}{l}{{\ul }}                                                               & \multicolumn{1}{l|}{}                                                         & \multicolumn{1}{l}{}                                                   & \multicolumn{1}{l}{}                                                                    & \multicolumn{1}{l}{}                                                   \\ \hline
    \hline
    
    \multicolumn{1}{c|}{\multirow{4}{*}{\textbf{Dx}}}    & CodeEmb    & 0.757                                                                       & {\ul{0.752}\text{**}}                                                                   & {0.755}                                                                & 0.562                                                                  & {0.558}                                                                          & {0.563}                                                         \\
    \multicolumn{1}{c|}{}                                & FT-BERT    & 0.771                                                                       & \textbf{0.775*}                                                                          & \textbf{0.777*}                                                               & 0.594                                                                  & \textbf{0.608**}                                                                        & \textbf{0.611*}                                                        \\
    \multicolumn{1}{c|}{}                                & SC-RNN     & 0.768                                                                       & 0.762                                                                                    & \textbf{0.773**}                                                              & 0.594                                                                  & \textbf{0.602**}                                                                        & {0.589}                                                         \\
    \multicolumn{1}{c|}{}                                & SC-RNN+MLM & 0.763                                                                       & 0.76                                                                                     & 0.768                                                                         & 0.583                                                                  & 0.586                                                                                   & \textbf{0.595*}                                                        \\ \hline
    \multicolumn{1}{c|}{\multirow{4}{*}{\textbf{Mort}}}  & CodeEmb    & 0.313                                                                       & 0.313                                                                                    & 0.313                                                                         & 0.24                                                                   & {0.233}                                                                          & {0.247}                                                         \\
    \multicolumn{1}{c|}{}                                & FT-BERT    & 0.378                                                                       & 0.378                                                                                    & 0.376                                                                         & 0.224                                                                  & \textbf{0.246*}                                                                         & \textbf{0.248*}                                                        \\
    \multicolumn{1}{c|}{}                                & SC-RNN     & 0.4                                                                         & 0.385                                                                                    & 0.401                                                                         & 0.252                                                                  & \textbf{0.267*}                                                                         & 0.252                                                                  \\
    \multicolumn{1}{c|}{}                                & SC-RNN+MLM & 0.393                                                                       & 0.383                                                                                    & 0.402                                                                         & 0.259                                                                  & 0.263                                                                                   & {0.253}                                                         \\ \hline
    \multicolumn{1}{c|}{\multirow{4}{*}{\textbf{LOS>3}}} & CodeEmb    & 0.61                                                                        & {0.606}                                                                           & 0.611                                                                         & 0.525                                                                  & {0.531}                                                                          & \textbf{0.534*}                                                        \\
    \multicolumn{1}{c|}{}                                & FT-BERT    & 0.624                                                                       & \textbf{0.628*}                                                                          & 0.624                                                                         & 0.536                                                                  & \textbf{0.542*}                                                                         & \textbf{0.549*}                                                        \\
    \multicolumn{1}{c|}{}                                & SC-RNN     & 0.634                                                                       & 0.632                                                                                    & {0.63}                                                                 & 0.54                                                                   & 0.543                                                                                   & \textbf{0.549*}                                                        \\
    \multicolumn{1}{c|}{}                                & SC-RNN+MLM & 0.628                                                                       & 0.627                                                                                    & \textbf{0.638*}                                                               & 0.537                                                                  & 0.541                                                                                   & \textbf{0.548*}                                                        \\ \hline
    \multicolumn{1}{c|}{\multirow{4}{*}{\textbf{LOS>7}}} & CodeEmb    & 0.326                                                                       & 0.333                                                                                    & 0.334                                                                         & 0.233                                                                  & {0.235}                                                                          & \textbf{0.239*}                                                        \\
    \multicolumn{1}{c|}{}                                & FT-BERT    & 0.36                                                                        & 0.356                                                                                    & 0.354                                                                         & 0.22                                                                   & \textbf{0.230*}                                                                         & \textbf{0.242**}                                                       \\
    \multicolumn{1}{c|}{}                                & SC-RNN     & 0.352                                                                       & 0.345                                                                                    & 0.35                                                                          & 0.229                                                                  & \textbf{0.236*}                                                                         & \textbf{0.253**}                                                       \\
    \multicolumn{1}{c|}{}                                & SC-RNN+MLM & 0.353                                                                       & 0.342                                                                                    & 0.342                                                                         & 0.234                                                                  & 0.235                                                                                   & \textbf{0.239*}                                                        \\ \hline
    \multicolumn{1}{c|}{\multirow{4}{*}{\textbf{ReAdm}}} & CodeEmb    & 0.043                                                                       & 0.044                                                                                    & 0.049                                                                         & 0.217                                                                  & {0.218}                                                                             & \textbf{0.232*}                                                        \\
    \multicolumn{1}{c|}{}                                & FT-BERT    & 0.043                                                                       & 0.044                                                                                    & 0.051                                                                         & 0.289                                                                  & {\ul{0.274}\text{*}}                                                                            & 0.281                                                                  \\
    \multicolumn{1}{c|}{}                                & SC-RNN     & 0.041                                                                       & 0.045                                                                                    & 0.046                                                                         & 0.28                                                                   & 0.263                                                                                   & {0.279}                                                         \\
    \multicolumn{1}{c|}{}                                & SC-RNN+MLM & 0.044                                                                       & 0.044                                                                                    & 0.044                                                                         & 0.255                                                                  & 0.255                                                                                   & \textbf{0.275*}                                                        \\ \hline
    \bottomrule
    \end{tabular}}
    \begin{flushleft}
    {\footnotesize* : p value < 0.05, \footnotesize** : p value < 0.01}
    \end{flushleft}
\end{table*}
\vspace{-3mm}
\textbf{Datasets\quad}
We draw on two publicly available datasets: MIMIC-III \citep{mimiciii} and eICU \citep{eicu}. MIMIC-III includes patients admitted to the intensive care unit (ICU) at Beth Israel Deaconess Medical Center, and eICU is a multi-center database across the United States. We extract laboratory measurements, medications prescribed, and infusion events from the first 12 hours of the first ICU stay of medical ICU (MICU) patients over the age of 18. As the result, we retain 18,536 samples from MIMIC-III and 12,818 samples from eICU, which are split into training, validation and testing sets according to a 65:15:20 ratio.

\noindent \textbf{Prediction Tasks\quad}
We examine a total of five prediction tasks based on individual ICU stays. Each task is defined as follows:
\vspace{-3mm}

\begin{enumerate}
 \item \textit{Readmission Prediction}:
Given a single ICU stay, we consider this sample readmitted if it is followed by another ICU stay during the same hospital stay.
\vspace{-3mm}

\item  \textit{Mortality Prediction}:
A sample is labeled positive for mortality if the discharge state was ``expired''.

\item  \textit{Length-of-Stay Prediction}:
There are two cases for length of stay (LOS) prediction; whether a given ICU stay lasted longer than 3 days, and whether it lasted longer than 7 days.

\item \textit{Diagnosis Prediction} (Multi-Label):
Given all diagnosis codes accumulated during the entire hospital stay, we group them into 18 diagnosis classes of the Clinical Classification Software (CCS) for ICD-9-CM criteria \citep{healthcare2016hcup}.
\end{enumerate}
\vspace{-5mm}
\noindent \textbf{Implementation Details\quad}
Based on the study for variant BERT models in Appendix~\ref{supp:domain_res}, we utilize BERT-Tiny architecture for the BERT-based text encoder.
Moreover, as shown in Appendix~\ref{supp:table}, we tested variant CodeEmb and DescEmb models to investigate the best architectures for our experimental settings. As the result, we finally choose randomly initialized code-based embedding model (CodeEmb), pre-trained BERT model (FT-BERT), RNN model (SC-RNN), and RNN model pre-trained on Masked Language Modeling (SC-RNN+MLM), for the further experiments.
Each experiment was evaluated using 10 random seeds, using area under precision recall curve (AUPRC) for binary classification tasks (readmission, mortality, LOS>3days, LOS>7days) and micro-averaged AUPRC (micro-AUPRC) for a multi-label classification task(diagnosis prediction). Evaluation metrics are reported throughout the paper as the mean of the ten seed experiments. We employ one fixed set of hyperparameters for all experiments; specific values are reported in Appendix~\ref{supp:hyperparameter}.


\begin{figure*}[ht]
  \centering
  \includegraphics[width=\linewidth]{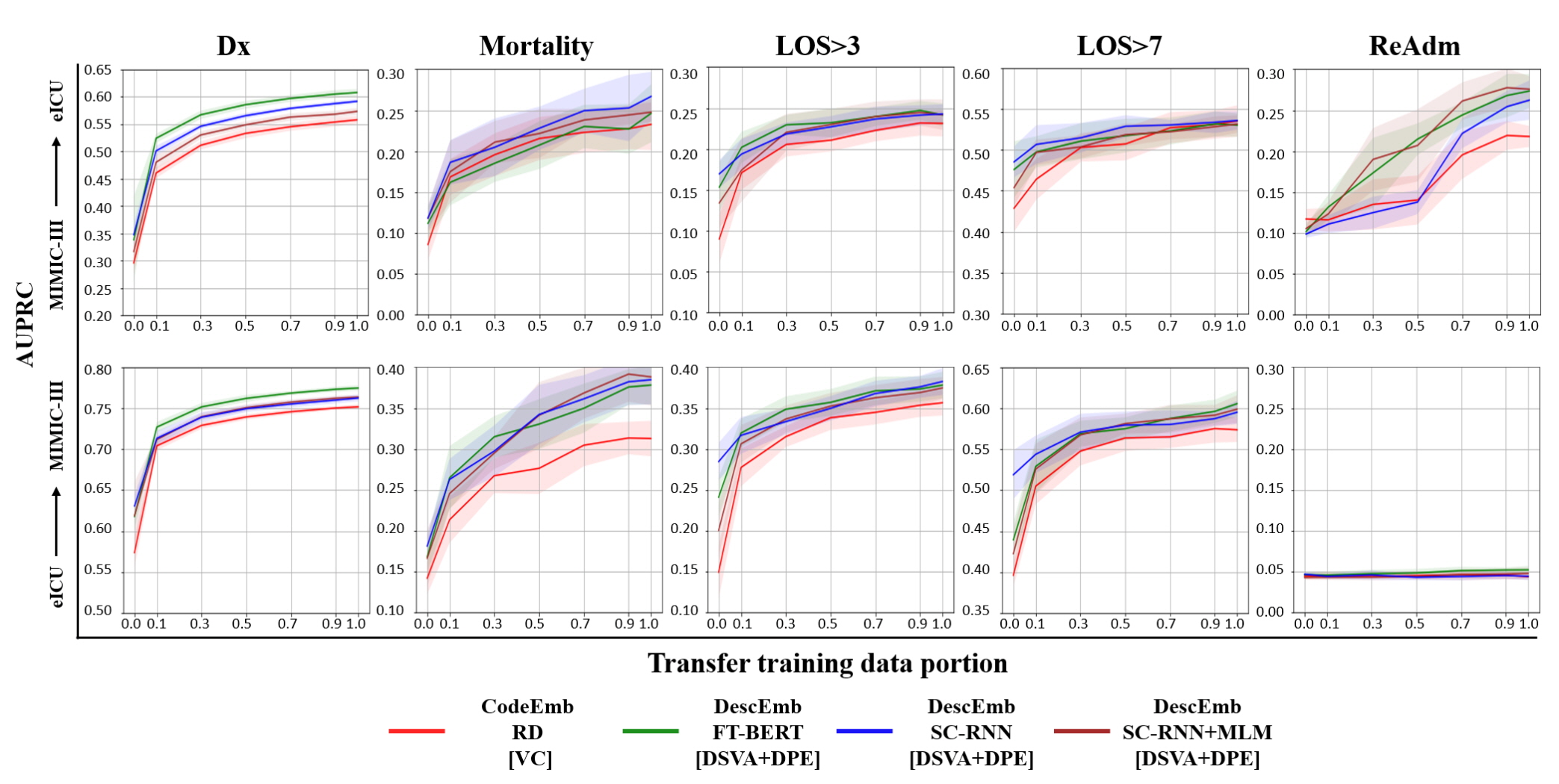}
  \caption{\label{fewshot} \textbf{Transfer learning performance (Top: MIMIC-III to eICU, Bottom: eICU to MIMIC-III).} The X-axis is the portion of the9 target dataset used for transfer learning, and the Y-axis is the AUPRC at test time on the target dataset. Shades represent the standard deviation from ten seed experiments.}
\end{figure*}
\vspace{-5mm}

\subsection{Prediction Performance}
\vspace{-2mm}
The "Single" columns in Table \ref{pooled} show the AUPRC of each prediction task on the two single datasets. Specifically, DescEmb models (FT-BERT, SC-RNN, SC-RNN+MLM) achieve comparable or superior performance to CodeEmb on nearly every case. 
\vspace{-5mm}
\subsection{Zero-Shot and Few-Shot Transfer}
\vspace{-2mm}
Because DescEmb’s embedding space is determined not by a code structure, our framework lends itself to transfer learning across all hospitals regardless of their EHR format. On the other hand, in order to deploy a code-based model on a target dataset with a different code structure, the new code embeddings received by the predictive layer must be randomly initialized, as $\textit{E}_\psi$ is not shared between hospitals. Here, we transfer one CodeEmb model and three DescEmb models trained on the MIMIC-III to eICU dataset and vice versa on zero shot and multiple few shot ratios. The results are shown in Figure \ref{fewshot}.

We observe higher performance of DescEmb over CodeEmb in all transfer setting. For all tasks except readmission prediction, DescEmb gains an advantage in zero-shot and smaller few-shot ratio transfer learnings, especially for the length-of-stay prediction tasks. This implies that DescEmb can be transferred to different hospitals while retaining its performance even for hospitals with a very small amount of data.
DescEmb can use its knowledge of a prior dataset to generate effective embeddings at the outset.

\vspace{-7mm}
\subsection{Pooled Learning with Distinct EHR Formats}
\vspace{-2mm}
If we were to deploy a large-scale predictive model in reality, it is more likely that a single central server would pool EHR data from multiple institutions and train a large-scale deep learning model, which we will further refer to as \textit{pooled learning}. Applying CodeEmb to pooled learning, however, requires substantial amount of time and labor. Conversely, given that DescEmb is not restricted by specific code structures, pooling datasets does not require any further preprocessing nor extra investment of time and money. 

In order to confirm the efficacy of DescEmb in the pooled learning scenario, we trained both DescEmb and CodeEmb on the pooled training set from both MIMIC-III and eICU, and tested on the individual test set. The results are reported in "Pooled" columns in Table \ref{pooled}. Compared to "Single" and "Transfer" which require individual model training on each dataset, "Pooled" has the operational efficiency since it only requires a single model training on the pooled dataset.

Within pooled learning, DescEmb outperformed CodeEmb in all cases (8.9\%P at most) except for readmission prediction for MIMIC-III. Of note, DescEmb’s pooled training showed favorable results compared to the single domain setting as well as transfer learning setting for both MIMIC-III and eICU.




\vspace{-7mm}
\section{Conclusion}
\vspace{-2mm}
In this work we introduced a code-agnostic predictive model for EHR, the description-based embedding (DescEmb), which unifies heterogeneous code systems by deriving the medical code embeddings with a neural text encoder. We demonstrated the superiority of DescEmb against CodeEmb in prediction, zero-shot and few-shot transfer learning and pooled learning. We believe this new framework will launch new discussion around large-scale model training for EHR. The code we used to train our models is available at https://github.com/hoon9405/DescEmb.

\acks{This work was supported by Institute of Information \& Communications Technology Planning \& Evaluation (IITP) grant (No.2019-0-00075, Artificial Intelligence Graduate School Program(KAIST)) funded by the Korea government (MSIT), and supported by the Korea Medical Device Development Fund grant funded by the Korea government (the Ministry of Science and ICT, the Ministry of Trade, Industry and Energy, the Ministry of Health \& Welfare, the Ministry of Food and Drug Safety) (Project Number: 1711138160, KMDF\_PR\_20200901\_0097).}

\bibliography{references}

\begin{thebibliography}{12}
\providecommand{\natexlab}[1]{#1}
\providecommand{\url}[1]{\texttt{#1}}
\expandafter\ifx\csname urlstyle\endcsname\relax
  \providecommand{\doi}[1]{doi: #1}\else
  \providecommand{\doi}{doi: \begingroup \urlstyle{rm}\Url}\fi

\bibitem[Che et~al.(2015)Che, Kale, Li, Bahadori, and Liu]{che2015deep}
Zhengping Che, David Kale, Wenzhe Li, Mohammad~Taha Bahadori, and Yan Liu.
\newblock Deep computational phenotyping.
\newblock In \emph{Proceedings of the 21th ACM SIGKDD International Conference
  on Knowledge Discovery and Data Mining}, pages 507--516, 2015.

\bibitem[Choi et~al.(2016{\natexlab{a}})Choi, Bahadori, Kulas, Schuetz,
  Stewart, and Sun]{choi2016retain}
Edward Choi, Mohammad~Taha Bahadori, Joshua~A Kulas, Andy Schuetz, Walter~F
  Stewart, and Jimeng Sun.
\newblock Retain: An interpretable predictive model for healthcare using
  reverse time attention mechanism.
\newblock \emph{arXiv preprint arXiv:1608.05745}, 2016{\natexlab{a}}.

\bibitem[Choi et~al.(2016{\natexlab{b}})Choi, Bahadori, Schuetz, Stewart, and
  Sun]{choi2016doctor}
Edward Choi, Mohammad~Taha Bahadori, Andy Schuetz, Walter~F Stewart, and Jimeng
  Sun.
\newblock Doctor ai: Predicting clinical events via recurrent neural networks.
\newblock In \emph{Machine learning for healthcare conference}, pages 301--318.
  PMLR, 2016{\natexlab{b}}.

\bibitem[Choi et~al.(2016{\natexlab{c}})Choi, Bahadori, Searles, Coffey,
  Thompson, Bost, Tejedor-Sojo, and Sun]{choi2016multi}
Edward Choi, Mohammad~Taha Bahadori, Elizabeth Searles, Catherine Coffey,
  Michael Thompson, James Bost, Javier Tejedor-Sojo, and Jimeng Sun.
\newblock Multi-layer representation learning for medical concepts.
\newblock In \emph{Proceedings of the 22nd ACM SIGKDD International Conference
  on Knowledge Discovery and Data Mining}, pages 1495--1504,
  2016{\natexlab{c}}.

\bibitem[Choi et~al.(2017)Choi, Bahadori, Song, Stewart, and Sun]{choi2017gram}
Edward Choi, Mohammad~Taha Bahadori, Le~Song, Walter~F Stewart, and Jimeng Sun.
\newblock Gram: graph-based attention model for healthcare representation
  learning.
\newblock In \emph{Proceedings of the 23rd ACM SIGKDD international conference
  on knowledge discovery and data mining}, pages 787--795, 2017.

\bibitem[Cost and (HCUP)(2016)]{healthcare2016hcup}
"Healthcare Cost and Utilization~Project" (HCUP).
\newblock Hcup clinical classifications software (ccs) for icd-9-cm, 2016.

\bibitem[Johnson et~al.(2016)Johnson, Pollard, Shen, Lehman, Feng, Ghassemi,
  Moody, Szolovits, Celi, and Mark]{mimiciii}
Alistair~EW Johnson, Tom~J Pollard, Lu~Shen, Li{-}wei~H Lehman, Mengling Feng,
  Mohammad Ghassemi, Benjamin Moody, Peter Szolovits, Leo~Anthony Celi, and
  Roger~G Mark.
\newblock Mimic-iii, a freely accessible critical care database.
\newblock \emph{Scientific data}, 3:\penalty0 160035, 2016.

\bibitem[Mikolov et~al.(2013)Mikolov, Chen, Corrado, and
  Dean]{mikolov2013efficient}
Tomas Mikolov, Kai Chen, Greg Corrado, and Jeffrey Dean.
\newblock Efficient estimation of word representations in vector space.
\newblock \emph{arXiv preprint arXiv:1301.3781}, 2013.

\bibitem[Miotto et~al.(2016)Miotto, Li, Kidd, and Dudley]{miotto2016deep}
Riccardo Miotto, Li~Li, Brian~A Kidd, and Joel~T Dudley.
\newblock Deep patient: an unsupervised representation to predict the future of
  patients from the electronic health records.
\newblock \emph{Scientific reports}, 6\penalty0 (1):\penalty0 1--10, 2016.

\bibitem[Pham et~al.(2016)Pham, Tran, Phung, and Venkatesh]{pham2016deepcare}
Trang Pham, Truyen Tran, Dinh Phung, and Svetha Venkatesh.
\newblock Deepcare: A deep dynamic memory model for predictive medicine.
\newblock In \emph{Pacific-Asia conference on knowledge discovery and data
  mining}, pages 30--41. Springer, 2016.

\bibitem[Pollard et~al.(2018)Pollard, Johnson, Raffa, Celi, Mark, and
  Badawi]{eicu}
Tom~J Pollard, Alistair~EW Johnson, Jesse~D Raffa, Leo~A Celi, Roger~G Mark,
  and Omar Badawi.
\newblock The eicu collaborative research database, a freely available
  multi-center database for critical care research.
\newblock \emph{Scientific data}, 5\penalty0 (1):\penalty0 1--13, 2018.

\bibitem[Rajkomar et~al.(2018)Rajkomar, Oren, Chen, Dai, Hajaj, Hardt, Liu,
  Liu, Marcus, Sun, et~al.]{rajkomar2018scalable}
Alvin Rajkomar, Eyal Oren, Kai Chen, Andrew~M Dai, Nissan Hajaj, Michaela
  Hardt, Peter~J Liu, Xiaobing Liu, Jake Marcus, Mimi Sun, et~al.
\newblock Scalable and accurate deep learning with electronic health records.
\newblock \emph{NPJ Digital Medicine}, 1\penalty0 (1):\penalty0 1--10, 2018.

\end{thebibliography}

\appendix
\newpage \onecolumn
\section{DescEmb overview}
\label{supp:fig}
\begin{figure*}[ht]
  \centering
  \includegraphics[width=\linewidth]{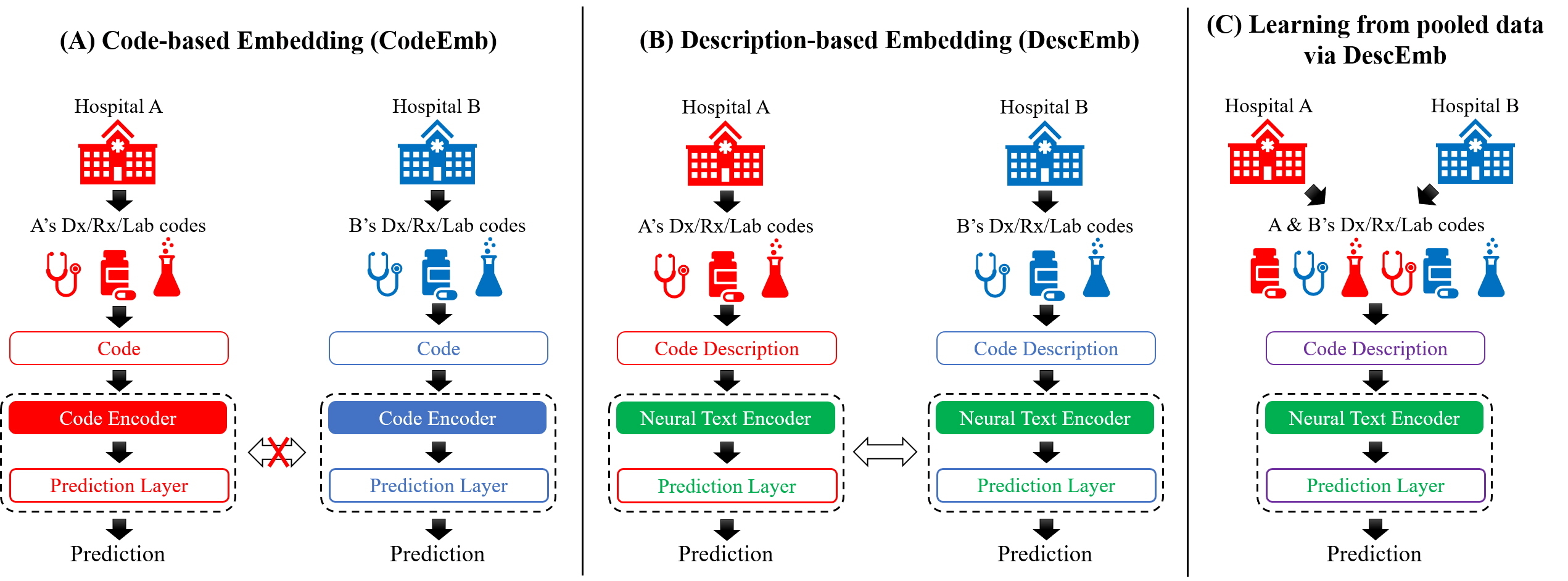}
  \caption{\label{Overview} \textbf{CodeEmb and DescEmb concept visualization} (A) CodeEmb: The code encoders and the prediction layers cannot be shared among different hospitals. (B) DescEmb: Both the text encoders and the prediction layers can be transferred between different hospitals. (C) Learning from pooled data via DescEmb: we can pool heterogeneous hospital data into one dataset and train jointly.}
\end{figure*}

\newpage

\section{AUPRC Results from Pre-Trained Text Encoders with Different Sizes and Pre-Training Techniques }
\label{supp:domain_res}
\begin{table*}[ht]
    \caption{\label{s1}\textbf{Results of BERT variation models on eICU}}
    \centering
    \resizebox{\textwidth}{!}{\begin{tabular}{cc|cccc|ccc}
    \toprule
    Task                                                 & Model   & BERT-tiny & BERT-mini & BERT-small & BERT  & Bio-BERT & Bio-clinical-BERT & Blue-BERT \\ \hline
    \hline
    
    \multicolumn{1}{c|}{\multirow{2}{*}{\textbf{Dx}}}    & CLS-FT  & 0.557     & 0.559     & 0.558      & 0.556 & 0.556    & 0.558             & 0.559     \\
    \multicolumn{1}{c|}{}                                & FT-BERT & 0.594     & 0.595     & 0.595      & 0.591 & 0.59     & 0.593             & 0.591     \\ \hline 
    \multicolumn{1}{c|}{\multirow{2}{*}{\textbf{Mort}}}  & CLS-FT  & 0.238     & 0.242     & 0.233      & 0.228 & 0.231    & 0.228             & 0.228     \\
    \multicolumn{1}{c|}{}                                & FT-BERT & 0.224     & 0.223     & 0.22       & 0.219 & 0.219    & 0.215             & 0.216     \\ \hline 
    \multicolumn{1}{c|}{\multirow{2}{*}{\textbf{LOS>3}}} & CLS-FT  & 0.528     & 0.528     & 0.526      & 0.524 & 0.527    & 0.525             & 0.526     \\
    \multicolumn{1}{c|}{}                                & FT-BERT & 0.536     & 0.527     & 0.523      & 0.523 & 0.522    & 0.528             & 0.526     \\ \hline 
    \multicolumn{1}{c|}{\multirow{2}{*}{\textbf{LOS>7}}} & CLS-FT  & 0.229     & 0.233     & 0.228      & 0.222 & 0.223    & 0.226             & 0.228     \\
    \multicolumn{1}{c|}{}                                & FT-BERT & 0.22      & 0.218     & 0.215      & 0.214 & 0.213    & 0.217             & 0.215     \\ \hline
    \multicolumn{1}{c|}{\multirow{2}{*}{\textbf{ReAdm}}} & CLS-FT  & 0.194     & 0.239     & 0.238      & 0.231 & 0.239    & 0.223             & 0.237     \\
    \multicolumn{1}{c|}{}                                & FT-BERT & 0.289     & 0.283     & 0.278      & 0.276 & 0.277    & 0.281             & 0.275     \\ \hline
    \end{tabular}}
\end{table*}
\begin{table*}[ht]
    \caption{\label{s2}\textbf{Results of BERT variation models on MIMIC-III}}
    \centering
    \resizebox{\textwidth}{!}{\begin{tabular}{cc|cccc|ccc}
    \toprule
    Task                                                 & Model   & BERT-tiny & BERT-mini & BERT-small & BERT  & Bio-BERT & Bio-clinical-BERT & Blue-BERT \\ \hline
    \hline
    
    \multicolumn{1}{c|}{\multirow{2}{*}{\textbf{Dx}}}    & CLS-FT  & 0.752     & 0.754     & 0.755      & 0.757 & 0.755    & 0.755             & 0.754     \\
    \multicolumn{1}{c|}{}                                & FT-BERT & 0.771     & 0.77      & 0.77       & 0.767 & 0.769    & 0.769             & 0.767     \\ \hline
    \multicolumn{1}{c|}{\multirow{2}{*}{\textbf{Mort}}}  & CLS-FT  & 0.339     & 0.345     & 0.34       & 0.344 & 0.339    & 0.338             & 0.335     \\
    \multicolumn{1}{c|}{}                                & FT-BERT & 0.378     & 0.371     & 0.365      & 0.362 & 0.363    & 0.363             & 0.364     \\ \hline
    \multicolumn{1}{c|}{\multirow{2}{*}{\textbf{LOS>3}}} & CLS-FT  & 0.616     & 0.614     & 0.615      & 0.615 & 0.611    & 0.61              & 0.608     \\
    \multicolumn{1}{c|}{}                                & FT-BERT & 0.624     & 0.623     & 0.623      & 0.621 & 0.626    & 0.622             & 0.62      \\ \hline
    \multicolumn{1}{c|}{\multirow{2}{*}{\textbf{LOS>7}}} & CLS-FT  & 0.346     & 0.344     & 0.341      & 0.343 & 0.344    & 0.344             & 0.338     \\
    \multicolumn{1}{c|}{}                                & FT-BERT & 0.36      & 0.352     & 0.342      & 0.342 & 0.345    & 0.345             & 0.343     \\ \hline
    \multicolumn{1}{c|}{\multirow{2}{*}{\textbf{ReAdm}}} & CLS-FT  & 0.044     & 0.043     & 0.044      & 0.045 & 0.044    & 0.045             & 0.044     \\
    \multicolumn{1}{c|}{}                                & FT-BERT & 0.043     & 0.043     & 0.044      & 0.043 & 0.045    & 0.043             & 0.043     \\ \hline
    \end{tabular}}
\end{table*}

Table \ref{s1} and Table \ref{s2} show AUPRC results differing in the size of pre-trained BERTs (BERT-tiny, BERT-mini, BERT-small, BERT) and in the domain-specific pre-training techniques (Bio-BERT, Bio-Clinical-BERT, Blue-BERT) in eICU and MIMIC-III respectively.  In this experiment, we tested CLS-FT and FT-BERT for verifying the effectiveness of the variants. From the table, there is no consistent performance tendency among different sizes of BERTs and pre-training techniques across tasks and models with very marginal performance differences. Of note, large text encoders generally underperform smaller sizes of BERT. Contrary to our expectation, domain-specialized pre-training techniques rather harm the model performance compared to smaller sizes of BERTs. Overall, the size of the text encoder influences the performance greatly more than how pre-training techniques are modeled. For the experiments in the main paper, we choose BERT-tiny since it generally shows decent performance among other models and it requires less memory and computation time compared to the large models. 

\newpage

\section{AUPRC of CodeEmbs and DescEmbs in prediction tasks with different value embedding methods and models}
\label{supp:table}
%

\begin{table*}[ht]
    \caption{\label{eicuAUPRC}\textbf{AUPRC of CodeEmbs and DescEmbs in prediction tasks for eICU}}
    \centering

    \resizebox{\textwidth}{!}{\begin{tabular}{cccccccccc}
    \toprule
    \multirow{2}{*}{}                                    & \multirow{2}{*}{Model}                    & \multicolumn{2}{c}{\multirow{2}{*}{CodeEmb}}        & \multicolumn{6}{c}{DescEmb}                                                                                                                                          \\ \cline{5-10} 
                                                         &                                      & \multicolumn{2}{c}{}                                & \multicolumn{4}{c}{BERT}                                                                            & \multicolumn{2}{c}{RNN}                                        \\ \hline
    \hline
    
    \multicolumn{1}{c|}{Task}                            & \multicolumn{1}{c|}{Value Embedding} & RD                    & \multicolumn{1}{c|}{W2V}    & CLS-FT & FT     & Scr   & \multicolumn{1}{c|}{\begin{tabular}[c]{@{}c@{}}FT \\  + MLM\end{tabular}} & Scr    & \begin{tabular}[c]{@{}c@{}}Scr \\  + MLM\end{tabular} \\ \hline
    \multicolumn{1}{c|}{\multirow{4}{*}{\textbf{Dx}}}    & \multicolumn{1}{c|}{VA}              & 0.447                 & \multicolumn{1}{c|}{0.433}  & 0.501  & 0.574  & 0.547 & \multicolumn{1}{c|}{0.586}                                                & 0.586  & 0.582                                                 \\
    \multicolumn{1}{c|}{}                                & \multicolumn{1}{c|}{DSVA}            & 0.447                 & \multicolumn{1}{c|}{0.433}  & 0.498  & 0.591  & 0.567 & \multicolumn{1}{c|}{0.601}                                                & 0.593  & 0.584                                                 \\
    \multicolumn{1}{c|}{}                                & \multicolumn{1}{c|}{DSVA+DPE}        & \multicolumn{1}{l}{\quad -} & \multicolumn{1}{l|}{\quad -}      & -      & 0.594  & 0.571 & \multicolumn{1}{c|}{0.602}                                                & 0.594  & 0.583                                                 \\
    \multicolumn{1}{c|}{}                                & \multicolumn{1}{c|}{VC}              & 0.562                 & \multicolumn{1}{c|}{0.549}  & 0.557  & 0.562  & 0.546 & \multicolumn{1}{c|}{0.555}                                                & 0.557  & 0.557                                                 \\ \hline
    \multicolumn{1}{c|}{\multirow{4}{*}{\textbf{Mort}}}  & \multicolumn{1}{c|}{VA}              & 0.112                 & \multicolumn{1}{c|}{0.153}  & 0.209  & 0.177  & 0.17  & \multicolumn{1}{c|}{0.216}                                                & 0.237† & 0.271                                                 \\
    \multicolumn{1}{c|}{}                                & \multicolumn{1}{c|}{DSVA}            & 0.112                 & \multicolumn{1}{c|}{0.153}  & 0.209  & 0.223  & 0.215 & \multicolumn{1}{c|}{0.213}                                                & 0.235  & 0.247                                                 \\
    \multicolumn{1}{c|}{}                                & \multicolumn{1}{c|}{DSVA+DPE}        & \multicolumn{1}{l}{\quad -} & \multicolumn{1}{l|}{\quad -}      & -      & 0.224  & 0.213 & \multicolumn{1}{c|}{0.217}                                                & 0.252  & 0.259                                                 \\
    \multicolumn{1}{c|}{}                                & \multicolumn{1}{c|}{VC}              & 0.24†                 & \multicolumn{1}{c|}{0.239†} & 0.238† & 0.23†  & 0.23† & \multicolumn{1}{c|}{0.223}                                                & 0.237† & 0.227†                                                \\ \hline
    \multicolumn{1}{c|}{\multirow{4}{*}{\textbf{LOS>3}}} & \multicolumn{1}{c|}{VA}              & 0.47†                 & \multicolumn{1}{c|}{0.439}  & 0.533  & 0.52   & 0.511 & \multicolumn{1}{c|}{0.514}                                                & 0.537  & 0.539                                                 \\
    \multicolumn{1}{c|}{}                                & \multicolumn{1}{c|}{DSVA}            & 0.47†                 & \multicolumn{1}{c|}{0.439}  & 0.529  & 0.53   & 0.538 & \multicolumn{1}{c|}{0.529}                                                & 0.539  & 0.537                                                 \\
    \multicolumn{1}{c|}{}                                & \multicolumn{1}{c|}{DSVA+DPE}        & \multicolumn{1}{l}{\quad-} & \multicolumn{1}{l|}{\quad -}      & -      & 0.536  & 0.537 & \multicolumn{1}{c|}{0.529}                                                & 0.54   & 0.537                                                 \\
    \multicolumn{1}{c|}{}                                & \multicolumn{1}{c|}{VC}              & 0.525                 & \multicolumn{1}{c|}{0.525}  & 0.528  & 0.523  & 0.524 & \multicolumn{1}{c|}{0.523}                                                & 0.526  & 0.53                                                  \\ \hline
    \multicolumn{1}{c|}{\multirow{4}{*}{\textbf{LOS>7}}} & \multicolumn{1}{c|}{VA}              & 0.157                 & \multicolumn{1}{c|}{0.184}  & 0.225  & 0.196† & 0.185 & \multicolumn{1}{c|}{0.196}                                                & 0.224  & 0.237                                                 \\
    \multicolumn{1}{c|}{}                                & \multicolumn{1}{c|}{DSVA}            & 0.157                 & \multicolumn{1}{c|}{0.184}  & 0.225  & 0.216  & 0.222 & \multicolumn{1}{c|}{0.221}                                                & 0.227  & 0.233                                                 \\
    \multicolumn{1}{c|}{}                                & \multicolumn{1}{c|}{DSVA+DPE}        & \multicolumn{1}{l}{\quad -} & \multicolumn{1}{l|}{\quad -}      & -      & 0.22   & 0.219 & \multicolumn{1}{c|}{0.221}                                                & 0.231  & 0.234                                                 \\
    \multicolumn{1}{c|}{}                                & \multicolumn{1}{c|}{VC}              & 0.231                 & \multicolumn{1}{c|}{0.228}  & 0.229  & 0.216  & 0.218 & \multicolumn{1}{c|}{0.218}                                                & 0.222  & 0.224                                                 \\ \hline
    \multicolumn{1}{c|}{\multirow{4}{*}{\textbf{ReAdm}}} & \multicolumn{1}{c|}{VA}              & 0.168                 & \multicolumn{1}{c|}{0.15}   & 0.208  & 0.283  & 0.205 & \multicolumn{1}{c|}{0.283}                                                & 0.269  & 0.279                                                 \\
    \multicolumn{1}{c|}{}                                & \multicolumn{1}{c|}{DSVA}            & 0.168                 & \multicolumn{1}{c|}{0.15}   & 0.206  & 0.284  & 0.264 & \multicolumn{1}{c|}{0.29}                                                 & 0.28   & 0.275                                                 \\
    \multicolumn{1}{c|}{}                                & \multicolumn{1}{c|}{DSVA+DPE}        & \multicolumn{1}{l}{\quad -} & \multicolumn{1}{l|}{\quad -}      & -      & 0.289† & 0.263 & \multicolumn{1}{c|}{0.284}                                                & 0.28   & 0.255                                                 \\
    \multicolumn{1}{c|}{}                                & \multicolumn{1}{c|}{VC}              & 0.217†                & \multicolumn{1}{c|}{0.183†} & 0.194  & 0.272  & 0.256 & \multicolumn{1}{c|}{0.267}                                                & 0.277  & 0.276                                                 \\ \hline
    \bottomrule
    \end{tabular}}
        \begin{flushleft}
        {\footnotesize †: standard deviation > 0.02}
        \end{flushleft}

\end{table*}

\begin{table*}[ht]
    \centering
    \caption{\label{mimicauprc} \textbf{AUPRC of CodeEmb and DescEmb in prediction tasks for MIMIC-III.}}
    \resizebox{\textwidth}{!}{\begin{tabular}{cccccccccc}
    \toprule
     \multirow{2}{*}{}                                    & \multirow{2}{*}{}                    & \multicolumn{2}{c}{\multirow{2}{*}{CodeEmb}} & \multicolumn{6}{c}{DescEmb}                                                                                                                      \\ \cline{5-10} 
                                                         &                                      & \multicolumn{2}{c}{}                         & \multicolumn{4}{c}{BERT}                                                        & \multicolumn{2}{c}{RNN}                                        \\ \hline
    \hline
   
    \multicolumn{1}{c|}{Task}                            & \multicolumn{1}{c|}{Value Embedding} & RD          & W2V                            & CLS-FT & FT     & Scr    & \begin{tabular}[c]{@{}c@{}}FT \\  + MLM\end{tabular} & Scr    & \begin{tabular}[c]{@{}c@{}}Scr \\  + MLM\end{tabular} \\ \hline
    \multicolumn{1}{c|}{\multirow{4}{*}{\textbf{Dx}}}    & \multicolumn{1}{c|}{VA}              & 0.726       & \multicolumn{1}{c|}{0.704}     & 0.733  & 0.76   & 0.747  & \multicolumn{1}{c|}{0.767}                           & 0.767  & 0.762                                                 \\
    \multicolumn{1}{c|}{}                                & \multicolumn{1}{c|}{DSVA}            & 0.726       & \multicolumn{1}{c|}{0.704}     & 0.731  & 0.77   & 0.752  & \multicolumn{1}{c|}{0.776}                           & 0.77   & 0.766                                                 \\
    \multicolumn{1}{c|}{}                                & \multicolumn{1}{c|}{DSVA+DPE}        & -           & \multicolumn{1}{c|}{-}         & -      & 0.771  & 0.752  & \multicolumn{1}{c|}{0.764}                           & 0.768  & 0.763                                                 \\
    \multicolumn{1}{c|}{}                                & \multicolumn{1}{c|}{VC}              & 0.757       & \multicolumn{1}{c|}{0.751}     & 0.752  & 0.756  & 0.745  & \multicolumn{1}{c|}{0.75}                            & 0.755  & 0.753                                                 \\ \hline
    \multicolumn{1}{c|}{\multirow{4}{*}{\textbf{Mort}}}  & \multicolumn{1}{c|}{VA}              & 0.228       & \multicolumn{1}{c|}{0.209}     & 0.346  & 0.343† & 0.31   & \multicolumn{1}{c|}{0.38}                            & 0.383  & 0.398                                                 \\
    \multicolumn{1}{c|}{}                                & \multicolumn{1}{c|}{DSVA}            & 0.228       & \multicolumn{1}{c|}{0.209}     & 0.347  & 0.377  & 0.378  & \multicolumn{1}{c|}{0.379}                           & 0.394† & 0.39                                                  \\
    \multicolumn{1}{c|}{}                                & \multicolumn{1}{c|}{DSVA+DPE}        & -           & \multicolumn{1}{c|}{-}         & -      & 0.378  & 0.372  & \multicolumn{1}{c|}{0.383}                           & 0.4    & 0.393                                                 \\
    \multicolumn{1}{c|}{}                                & \multicolumn{1}{c|}{VC}              & 0.313†      & \multicolumn{1}{c|}{0.334}     & 0.339  & 0.336† & 0.335† & \multicolumn{1}{c|}{0.376}                           & 0.344† & 0.338                                                 \\ \hline
    \multicolumn{1}{c|}{\multirow{4}{*}{\textbf{LOS>3}}} & \multicolumn{1}{c|}{VA}              & 0.582       & \multicolumn{1}{c|}{0.585}     & 0.608  & 0.616  & 0.601  & \multicolumn{1}{c|}{0.616}                           & 0.624  & 0.63                                                  \\
    \multicolumn{1}{c|}{}                                & \multicolumn{1}{c|}{DSVA}            & 0.582       & \multicolumn{1}{c|}{0.585}     & 0.608  & 0.624  & 0.617  & \multicolumn{1}{c|}{0.619}                           & 0.631  & 0.632                                                 \\
    \multicolumn{1}{c|}{}                                & \multicolumn{1}{c|}{DSVA +DPE}       & -           & \multicolumn{1}{c|}{-}         & -      & 0.624  & 0.616  & \multicolumn{1}{c|}{0.622}                           & 0.634  & 0.628                                                 \\
    \multicolumn{1}{c|}{}                                & \multicolumn{1}{c|}{VC}              & 0.61        & \multicolumn{1}{c|}{0.614}     & 0.616  & 0.61   & 0.614  & \multicolumn{1}{c|}{0.612}                           & 0.622  & 0.622                                                 \\ \hline
    \multicolumn{1}{c|}{\multirow{4}{*}{\textbf{LOS>7}}} & \multicolumn{1}{c|}{VA}              & 0.269       & \multicolumn{1}{c|}{0.251}     & 0.346  & 0.338  & 0.325  & \multicolumn{1}{c|}{0.342}                           & 0.349  & 0.349                                                 \\
    \multicolumn{1}{c|}{}                                & \multicolumn{1}{c|}{DSVA}            & 0.269       & \multicolumn{1}{c|}{0.251}     & 0.348  & 0.355  & 0.359  & \multicolumn{1}{c|}{0.356}                           & 0.35   & 0.35                                                  \\
    \multicolumn{1}{c|}{}                                & \multicolumn{1}{c|}{DSVA+DPE}        & -           & \multicolumn{1}{c|}{-}         & -      & 0.36   & 0.359  & \multicolumn{1}{c|}{0.353}                           & 0.352  & 0.353                                                 \\
    \multicolumn{1}{c|}{}                                & \multicolumn{1}{c|}{VC}              & 0.326       & \multicolumn{1}{c|}{0.342}     & 0.346  & 0.341  & 0.339  & \multicolumn{1}{c|}{0.344}                           & 0.347  & 0.352                                                 \\ \hline
    \multicolumn{1}{c|}{\multirow{4}{*}{\textbf{ReAdm}}} & \multicolumn{1}{c|}{VA}              & 0.044       & \multicolumn{1}{c|}{0.043}     & 0.042  & 0.042  & 0.045† & \multicolumn{1}{c|}{0.044}                           & 0.044  & 0.043                                                 \\
    \multicolumn{1}{c|}{}                                & \multicolumn{1}{c|}{DSVA}            & 0.044       & \multicolumn{1}{c|}{0.043}     & 0.041  & 0.043  & 0.046† & \multicolumn{1}{c|}{0.044}                           & 0.045  & 0.044                                                 \\
    \multicolumn{1}{c|}{}                                & \multicolumn{1}{c|}{DSVA+DPE}        & -           & \multicolumn{1}{c|}{-}         & -      & 0.043  & 0.047  & \multicolumn{1}{c|}{0.044}                           & 0.041  & 0.044                                                 \\
    \multicolumn{1}{c|}{}                                & \multicolumn{1}{c|}{VC}              & 0.043       & \multicolumn{1}{c|}{0.043}     & 0.044  & 0.045  & 0.047  & \multicolumn{1}{c|}{0.044}                           & 0.044  & 0.044                                                 \\ \hline
    \bottomrule
    \end{tabular}}
        \begin{flushleft}
        {\footnotesize †: standard deviation > 0.02}
        \end{flushleft}
        
\end{table*}

\subsection{Abbreviation explanation}
For value embedding, Value Aggregation (VA) stands for aggregating the code description and the numeric values together as text. In this setting, because the BERT tokenizer recognizes each value as a word, it sometimes tokenizes a given value in an unnatural way. For example, a number ‘1351’ can be split into two sub-words ‘13’ and ‘51’, which does not best reflect the underlying meaning of the number. Hence, we additionally propose Digit-Split Value Aggregation (DSVA), where we split all numeric values into each digit first, then aggregate with the code description as text. In this way, a number is always tokenized into single digits, and also we add learnable Digit Place Embedding (DPE) to every digit token indicating its place value, named Digit-Split Value Aggregation + Digit Place Embedding (DSVA+DPE). This can only be applied to the model exploiting a neural text encoder which can add additional value embedding for each digit, so the results for DSVA + DPE in CodeEmb and CLS-FT are blank since they cannot use Digit Place Embedding. Value embedding are abbreviated as VA, DSVA, DSVA+DPE in order. In CodeEmb, ‘RD’ represents a randomly initialized embedding layer while ‘W2V’ represents Word2Vec, a pre-training strategy for CodeEmb embedding layer \cite{mikolov2013efficient}. ‘FT’ stands for fine-tuning where we employ existing pre-trained BERT parameters and fine-tune them for the downstream tasks. ‘Scr’ stands for training from scratch where we do not bring the pre-trained BERT but randomly initialize the model. ‘FT + MLM’ is a model that brings a pre-trained model and conducts Masked Language Modeling (MLM) on our dataset after which it is fine-tuned on downstream tasks. ‘Scr + MLM’ is similar to ‘FT + MLM’ but it does not bring the pre-trained model parameter. We utilize the BERT-Tiny architecture for the BERT-based text encoder

\subsection{CodeEmb vs DescEmb performance result}
DescEmb models achieve comparable or superior performance to CodeEmb on nearly every task across all value embedding methods at the average of 8\%P with 12\%P at maximum. Within DescEmb models, BERT-FT generally outperforms BERT-Scratch, verifying the effectiveness of pre-training on massive text corpus. Using the additional Masked Language Modeling (MLM) on our dataset marginally improved performance (+0.3\%P AUPRC) for BERT models. We further test the efficacy of MLM in the transfer learning setting below. Of note, a Bi-RNN text encoder generally performs better than BERT-based models. We speculate that, since the maximum lengths of sub-tokens for one code description are 46 and 48 for MIMIC-III and eICU respectively, a simple and light-weighted text encoder model, in this case Bi-RNN, has enough capacity to grasp the features of descriptions. In other words, a large and complex model, in this case BERT, might be an excessively powerful tool to compute refined representations in our setting.

Note that CLS-finetune in DescEmb, which requires the same amount of compute and time as CodeEmb but initializes the embeddings with the CLS outputs from pre-trained BERT, outperforms CodeEmb in nearly all cases. This demonstrates that there is ground to be gained by adopting description-based embedding compared to the classical code-based embedding. We also pre-train the CodeEmb’s embedding layer in Word2Vec manner to have a fair comparison with the pre-trained DescEmb models. We observe that Word2Vec results are highly unstable, which sometimes underperform 3.4\%P at the worst compared to randomly initialized CodeEmb. This result implies that pre-training at code-level is insufficient to fully capture the semantics of each code and sometimes harms the performance. On the other hand, all pre-trained DescEmb models consistently show high performance across all scenarios verifying the robustness of pre-training at description-level. 

For value embeddings, there is a large discrepancy between CodeEmbs and DescEmbs in VA and DSVA compared to other value embedding methods. We conjecture the underlying reason is that in VA and DSVA, the unique number of codes for CodeEmb explodes since a new code is needed when different values are used.  This raises the curse of dimensionality.  On the contrary, the unique number of sub-tokens used in DescEmb does not change significantly in either setting, resulting in a stable performance. Hence, DescEmb is a suitable model architecture for understanding values because it does not require creating a new code for different values. VC performs the best in CodeEmb. In DescEmb, Digit-Split Value Aggregation with DSVA+DPE shows higher performance on the whole than other value embedding methods. It explains that the model has better numeric understanding since DPE explicitly notifies the model about the place value. For further experiments, we choose CodeEmb RD, FT-BERT, SC-RNN, SC-RNN + MLM, with VC for CodeEmb and DSVA+DPE for the DescEmb models.

\newpage 
\section{PCA Results for Varied Random Seeds}

\begin{figure*}[h]
  \centering
  \includegraphics[width=\linewidth]{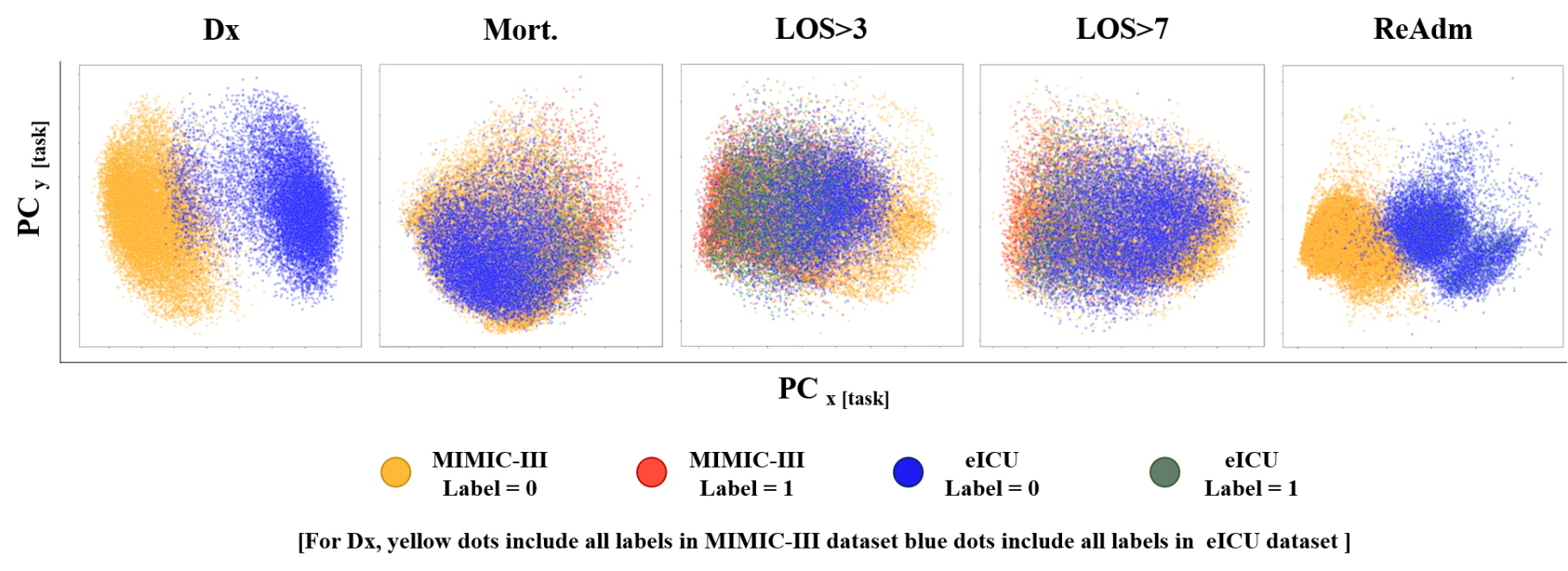}
  \caption{\label{pca} \textbf{PCA visualizations of the ICU representations from the two datasets.} The X-axis and Y-axis correspond to two different principal components. Each dot represents one ICU stay, and the dot color represents the target label for binary classification tasks. For the diagnosis prediction task (Dx), labels could not be succinctly annotated due to its multi-label classification nature. Thus, we distinguish dataset sources by color: yellow being MIMIC-III and blue being eICU.}
\end{figure*}

From Table \ref{pooled}, we can see that eICU generally gained more performance increase than MIMIC-III from both pooled learning and transfer learning. We hypothesize that this comes from the data distribution properties of the two datasets. In order to confirm our hypothesis, we conducted Principal Component Analysis (PCA) on the ICU stay representation vectors obtained from the prediction model (the last hidden layer of the RNN) trained on the pooled dataset. The results in Figure \ref{pca} show that, for some tasks, the eICU representations are distributed inside the MIMIC-III representation distributions, especially in LOS tasks where eICU gained notable performance increase from transfer and pooled learning compared to the single-domain learning.  We deduce that the performance increase comes from learning a more generally distributed dataset, in this case MIMIC-III.

\newpage 

\section{Detailed preprocessing method and table statistics}
\label{supp:preprocess}
\subsection{Detailed Preprocessing Information}
\label{supp:preprocess_stat}
In the following section we provide further detail about the construction of our datasets. As input for our predictive models, we employ three sources of information (we will further denote source of information as ‘item’ for simplicity)— laboratory, medication, and infusion—simultaneously for each patient. The .csv files corresponding to each item are described in Table \ref{filesource}. Note that when merging MIMIC-III files 'INPUTEVENTS\_MV’ and 'INPUTEVENTS\_CV’, we remove 41 patient histories which straddle the transition between code systems and consequently are included partially in each file.

\begin{table}[ht]
    \caption{\label{filesource} \textbf{File sources for each dataset}}
    \centering
    \begin{tabular}{l c c }
    \toprule
    Item & Source & Filename \\
    \hline
    \midrule
    Lab &  MIMIC-III  & LABEVENTS.csv  \\
    Lab &  eICU   & lab.csv        \\
    Med &  MIMIC-III  & PRESCRIPTIONS.csv\\
    Med &  eICU   & medication.csv \\
    Inf &  MIMIC-III  & INPUTEVENTS.csv\\
    Inf &  eICU   & infusionDrug.csv \\
    \bottomrule
    \end{tabular}
    
\end{table}

For the sake of comparability, we built patients cohorts from the full MIMIC-III and eICU databases based on the following criteria: (1) Medical ICU (MICU) patients (2) over the age of 18 who (3) remain in the ICU for over 12 hours. We operationalize criterion (1) in MIMIC-III as patients for whom the first care unit is the last care unit and ICU type is MICU (i.e. we exclude patients who have transferred ICUs). For patients with multiple ICU stays, we draw exclusively on the first stay, and we remove any ICU stays with fewer than 5 observed codes. Within each ICU stay, we restrict our sample to the first 150 codes during the first 12 hours of data, and remove codes which occur fewer than 5 times in the entire dataset. Code sequence is determined by the associated time stamp.

\begin{table}[ht]
    \caption{\label{predstats} \textbf{Prediction dataset summary statistics}}
    \centering
    \begin{tabular}{l l l }
    \toprule
    Statistic          &  eICU       &  MIMIC-III  \\
    \hline
    \midrule
    $N$ Observations   &  12,818     &    18,536                \\
    $N$ ICU Stays      &  12,818     &    18,536                \\
    $N$ Hospital Adm.  &  12,818     &    18,536                \\
    $N$ Patients       &  12,818     &    18,536                \\
    Mean Seq. Length   &  48.8       &    65.3                  \\
    Median Seq. Length &  43.0       &    57.0                  \\
    $N$ Total Codes    &  625,594    &    1,211,107             \\
    $N$ Unique Codes   &  2,018      &    2,855                 \\
    \bottomrule
    \end{tabular}
\end{table}

\subsection{Predictive Task Labels}
\label{supp:preprocess_label}
We predict patient outcomes across five tasks: readmission, mortality, an ICU stay exceeding three or seven days, and diagnosis prediction. The first four are binary classification, the last multi-label. The variable-level criteria to generate these labels is available in Table \ref{labelcriteria}.

In order to generate diagnosis labels for comparison across datasets, we employ the Clinical Classifications Software (CCS) for ICD-9-CM of the Healthcare Cost and Utilization Project \cite{healthcare2016hcup}. We utilize the highest level representation available of ICD9 diagnosis, a common code format across EHR. There are 18 such representations. MIMIC-III and eICU diagnoses represented by ICD9 codes are simply mapped using the CCS classification. eICU ICD10 diagnoses are mapped first to ICD9 codes before to their CCS classification. Finally, for eICU string diagnoses (e.g. Infection ... | ... bacterial ... | ... tuberculosis), we first search the most granular level for a string match with ICD9 before proceeding up the hierarchy for a match.

\begin{table*}[ht]
    \caption{\label{labelcriteria} \textbf{Specific label criteria}}
    
    \centering
    \begin{tabular}{l l l}
    \toprule
    Target                  & eICU                                       & MIMIC-III             \\
    \hline
    \midrule
    Readmission             & Count(‘patientUnitStayID') \textgreater 1  & Count(‘ICUSTAY\_ID') \textgreater 1  \\
    Mortality               & ‘unitDischargeStatus'==‘Expired'           & ‘DOD\_HOSP' not null                  \\
    LOS \textgreater 3 Days & ‘unitDischargeOffset' \textgreater 3*24*60 & LOS \textgreater 3                  \\
    LOS \textgreater 7 Days & ‘unitDischargeOffset' \textgreater 7*24*60 & LOS \textgreater 7                  \\
    Diagnosis               & set(‘diagnosisstring’) per 1 ICU           & ICD9\_CODE-LONG\_TITLE     \\
    \bottomrule
    \end{tabular}
\end{table*}

\subsection{Data Statistics}
After preprocessing input data, we found that some patients lack all three items. Consequently, in some cases the item was left out from the patient dataset. For example, some patients have all the items in the code sequence, while others are included without all of them.  In the MIMIC-III and eICU we use, the size of the entire dataset is the same as the union shown in Table \ref{predstats} for each of the source dataset.

\subsection{Hyperparameters}
\label{supp:hyperparameter}
We conducted the hyperparameter searching experiment in CodeEmb and DescEmb on MIMIC-III and eICU. We swept the hyperparameter space within a fixed range, presented below, by grid search.
\begin{itemize}
    \item dropout = [0.1, 0.3, 0.5]
    \item embedding dimension = [128, 256, 512, 768]
    \item hidden dimension = [128, 256, 512]
    \item learning rate = [5e-4, 1e-4, 5e-5, 1e-5]
\end{itemize}
We spent over 72 hours trying to find the best hyperparameter set for each case. We noticed that hyperparameters did not significantly affect the final result. For the experiment’s simplicity, we unified one hyperparameter set for all cases without greatly harming each individual model’s performance. The final set results are dropout of 0.3, embedding dimension and hidden dimension for the predictive model as 128 and 256 respectively, and learning rate of 1e-4.
\\
\\

\newpage
\section{Standard error for the results in the three scenarios: single domain learning, transfer learning, pooled}
To show the reliability of our results, the standard error for Table 1 is shown here. Through representing standard error, a more clear interpretation and robustness of the results can be achieved.
In the experiment, the number of each experiment conducted was 10 in all results, and there is no big difference in error for each model. However, it can be confirmed that the error is different depending on tasks, training data, and train and test settings.
As in Table1, CodeEmb (RD) value embedding type is VC, and DescEmb (FT-BERT , SC-RNN, SC-RNN + MLM) is DSVA+DPE. \\

\begin{table*}[ht]
    \centering
    \caption{\label{mimicauprc} \textbf{standard error for the models on the five prediction tasks in thethree scenarios: single domain learning, transfer learning, pooled}}
    \begin{tabular}{cc|ccc|ccc}

\hline
                                                                  &                        & Single    & Transfer    & Pooled    & Single & Transfer  & Pooled \\
\multirow{2}{*}{Task}                                             & \multirow{2}{*}{Model} & MIMIC-III & eICU        & MIMIC-III & eICU   & MIMIC-III & eICU   \\
                                                                  &                        &           & → MIMIC-III &           &        & → eICU    &        \\ \hline
                                                                  \hline
\multicolumn{1}{c|}{\multirow{4}{*}{\textbf{DX}}}                 & CodeEmb                & 0.0009    & 0.0022      & 0.0009    & 0.0025 & 0.0009    & 0.0022 \\
\multicolumn{1}{c|}{}                                             & FT-BERT                & 0.0009    & 0.0022      & 0.0009    & 0.0019 & 0.0009    & 0.0019 \\
\multicolumn{1}{c|}{}                                             & SC-RNN                 & 0.0009    & 0.0013      & 0.0013    & 0.0016 & 0.0013    & 0.0022 \\
\multicolumn{1}{c|}{}                                             & SC-RNN + MLM           & 0.0006    & 0.0016      & 0.0009    & 0.0019 & 0.0009    & 0.0019 \\ \hline
\multicolumn{1}{c|}{\multirow{4}{*}{\textbf{Mort}}}               & CodeEmb                & 0.0079    & 0.0101      & 0.0073    & 0.0104 & 0.007     & 0.0089 \\
\multicolumn{1}{c|}{}                                             & FT-BERT                & 0.007     & 0.0111      & 0.0079    & 0.0079 & 0.0076    & 0.0085 \\
\multicolumn{1}{c|}{}                                             & SC-RNN                 & 0.0092    & 0.0095      & 0.0095    & 0.0092 & 0.0095    & 0.0104 \\
\multicolumn{1}{c|}{}                                             & SC-RNN + MLM           & 0.0076    & 0.0104      & 0.0085    & 0.0089 & 0.0079    & 0.0089 \\ \hline
\multicolumn{1}{c|}{\multirow{4}{*}{\textbf{LOS\textgreater{}3}}} & CodeEmb                & 0.0054    & 0.0051      & 0.0054    & 0.0051 & 0.0051    & 0.0057 \\
\multicolumn{1}{c|}{}                                             & FT-BERT                & 0.0035    & 0.0044      & 0.0054    & 0.0057 & 0.0051    & 0.0054 \\
\multicolumn{1}{c|}{}                                             & SC-RNN                 & 0.006     & 0.0038      & 0.0047    & 0.0051 & 0.0051    & 0.0044 \\
\multicolumn{1}{c|}{}                                             & SC-RNN + MLM           & 0.0047    & 0.0047      & 0.0047    & 0.0054 & 0.0047    & 0.0047 \\ \hline
\multicolumn{1}{c|}{\multirow{4}{*}{\textbf{LOS\textgreater{}7}}} & CodeEmb                & 0.0063    & 0.006       & 0.006     & 0.0032 & 0.0047    & 0.0041 \\
\multicolumn{1}{c|}{}                                             & FT-BERT                & 0.0063    & 0.0044      & 0.0066    & 0.0035 & 0.0051    & 0.0035 \\
\multicolumn{1}{c|}{}                                             & SC-RNN                 & 0.0041    & 0.0035      & 0.0041    & 0.0028 & 0.0044    & 0.0047 \\
\multicolumn{1}{c|}{}                                             & SC-RNN + MLM           & 0.0057    & 0.0047      & 0.0057    & 0.0035 & 0.0041    & 0.0047 \\ \hline
\multicolumn{1}{c|}{\multirow{4}{*}{\textbf{ReAdm}}}              & CodeEmb                & 0.0006    & 0.0041      & 0.0016    & 0.0104 & 0.0013    & 0.0057 \\
\multicolumn{1}{c|}{}                                             & FT-BERT                & 0.0009    & 0.0066      & 0.0013    & 0.0073 & 0.0016    & 0.0054 \\
\multicolumn{1}{c|}{}                                             & SC-RNN                 & 0.0006    & 0.0076      & 0.0016    & 0.0066 & 0.0016    & 0.0076 \\
\multicolumn{1}{c|}{}                                             & SC-RNN + MLM           & 0.0006    & 0.0063      & 0.0016    & 0.0076 & 0.0013    & 0.0082 \\ \hline
\bottomrule
\end{tabular}
\end{table*}




\end{document}